\documentclass[conference]{IEEEtran}
\IEEEoverridecommandlockouts
\usepackage{cite}
\usepackage{amsmath,amssymb,amsfonts}
\usepackage{algorithmic}
\usepackage[ruled,vlined]{algorithm2e}
\usepackage{graphicx}
\usepackage{textcomp}
\usepackage{xcolor}
\usepackage{lipsum}
\usepackage{graphicx}
\usepackage{comment}
\usepackage{amsmath,amssymb} 
\usepackage{color}
\usepackage{cite}
\usepackage{multirow}
\usepackage{wrapfig,lipsum,booktabs}
\usepackage{makecell}
\usepackage{tabularx}
\usepackage{float}
\usepackage{stfloats}
\usepackage{tabu}
\usepackage[a4paper, total={184mm,239mm}]{geometry}

\newcommand\blfootnote[1]{%
  \begingroup
  \renewcommand\thefootnote{}\footnote{#1}%
  \addtocounter{footnote}{-1}%
  \endgroup
}

\mathchardef\simsym"0218
\def\BibTeX{{\rm B\kern-.05em{\sc i\kern-.025em b}\kern-.08em
    T\kern-.1667em\lower.7ex\hbox{E}\kern-.125emX}}
    
\begin{document}

\title{Activation Density based Mixed-Precision Quantization for Energy Efficient Neural Networks
}

\makeatletter
\newcommand{\linebreakand}{%
  \end{@IEEEauthorhalign}
  \hfill\mbox{}\par
  \mbox{}\hfill\begin{@IEEEauthorhalign}
}
\makeatother
\author{Karina Vasquez$^{*,1}$\thanks{\hspace{-3mm}$^*$ These authors have contributed equally to this work. This work was done while Karina was interning at Yale University.}, Yeshwanth Venkatesha$^{*,2}$, Abhiroop Bhattacharjee$^{2}$, Abhishek Moitra$^{2}$, and Priyadarshini Panda$^{2}$ 
 \\ $^{1}$Department of Electrical Engineering, UTEC, Peru \\
$^{2}$Department of Electrical Engineering, Yale University, USA 
\\ karina.vasquez@utec.edu.pe, \{yeshwanth.venkatesha, abhiroop.bhattacharjee, abhishek.moitra, priya.panda\}@yale.edu
}

\maketitle

\begin{abstract}
As neural networks gain widespread adoption in embedded devices, there is a growing need for model compression techniques to facilitate seamless deployment in resource-constrained environments. Quantization is one of the go-to methods yielding state-of-the-art model compression. Most quantization approaches take a fully trained model, then apply different heuristics to determine the optimal bit-precision for different layers of the network, and finally retrain the network to regain any drop in accuracy. Based on Activation Density—the proportion of non-zero activations in a layer—we propose a novel in-training quantization method. Our method calculates optimal bit-width/precision for each layer during training yielding an energy-efficient mixed precision model with competitive accuracy. Since we train lower precision models progressively during training, our approach yields the final quantized model at lower training complexity and also eliminates the need for re-training. We run experiments on benchmark datasets like CIFAR-10, CIFAR-100, TinyImagenet on VGG19/ResNet18 architectures and report the accuracy and energy estimates for the same. We achieve up to $4.5\times$ benefit in terms of estimated multiply-and-accumulate (MAC) reduction while reducing the training complexity by $50\%$ in our experiments. To further evaluate the energy benefits of our proposed method, we develop a mixed-precision scalable Process In Memory (PIM) hardware accelerator platform. The hardware platform incorporates shift-add functionality for handling multi-bit precision neural network models. Evaluating the quantized models obtained with our proposed method on the PIM platform yields about $5\times$ energy reduction compared to baseline 16-bit models. Additionally, we find that integrating activation density based quantization with activation density based pruning (both conducted during training) yields up to $\simsym198\times$ and $\simsym44\times$ energy reductions for VGG19 and ResNet18 architectures respectively on PIM platform compared to baseline 16-bit precision, unpruned models.
\end{abstract}


\section{Introduction}
\label{sec:intro}
\blfootnote{Published in Design, Automation and Test in Europe (DATE) conference.}
Neural network model compression has gained  popularity in the past few years as deep learning is being deployed in numerous applications across resource-constrained embedded devices \cite{embedded_device_dl}. As neural networks are known to have extremely high computation and memory requirements, there is a need for efficient processing algorithms in order to deploy them in energy constrained environments. \textit{Pruning}---removing the redundant parameters---and \textit{quantization}---using a lower bit representation for  the model parameters---are the two most popular techniques used to compress a model and significantly reduce the energy consumption \cite{han2015deep, model_compression_survey, energyawarepruning}.

The idea of quantized neural networks has been explored as early as the 1990s mainly to enable simpler hardware implementation \cite{fiesler1990weight, balzer1991weight, quantization_survey}. The complexity of quantization methods range from simple rounding \cite{courbariaux2015binaryconnect} to using deep reinforcement learning agents to optimize design decisions \cite{elthakeb2018releq}. In addition to quantizing weights, quantizing activations has also been found to be useful. The authors in \cite{mishra2017wrpn} report that activations consume more memory than weights. Furthermore, gradients can also be quantized which enables communication efficient training in a distributed learning system such as federated learning \cite{qsgd,fl_google}. Similar to quantization, network pruning also was introduced in the 1990s \cite{lecun1990optimal, reed1993pruning} and has garnered attention in the recent times due to the rapid deployment of neural networks on embedded platforms\cite{model_compression_survey}. 

A majority of pruning and quantization methods require a fully trained model as the starting point. Then, the model is pruned/quantized based on some heuristics followed by retraining to regain the drop in accuracy \cite{han2015deep}. Often this process of pruning/quantizing and retraining is repeated iteratively to obtain the final model. Note, only those quantization methods that are geared towards yielding mixed-precision networks incur such iterative process. Binarized  or homogeneous precision network implementations obtained by training models with same bit-width across all layers from scratch has been shown \cite{qnn}. However, such models generally suffer from accuracy loss as compared to mixed-precision models. But, in the latter, the prerequisite of a large fully trained network as a starting point is a significant overhead in the overall training-compression-retraining process. 

Generally, having a pre-trained model helps in assigning importance to the weights \cite{zhu2017prune}. However, recent evidence suggests that there is very little effect of having a fully trained model on the end-accuracy of the final pruned/quantized model. The authors in \cite{frankle2018lottery} showed that training a sparse model from randomly initialized weights resulted in similarly performing models as a baseline unpruned model. 
Leveraging this, we propose an in-training quantization method based on a novel activation density metric that yields a mixed-precision network and eliminates the need for a fully pre-trained model.

To demonstrate the energy and compute efficiency improvement of our proposed method, we design a precision-scalable Process In-Memory (PIM) hardware platform that can support variable data-precision and pruning for different layers. Implementation of multibit-precision functionality is relatively more viable in case of a PIM architecture since changing the bit-precisions does not lead to significant architectural changes as opposed to a conventional digital CMOS architecture. 

To cater to higher scalability and realistic mixed-precision implementations, we design our architecture to support only 2-/4-/8-/16-bit precisions. Thus, data precision of 3-bits would be translated to 4-bits, 5-bits to 8-bits, and so on. On the other hand, analytical methods based on a simple digital block capable of handling variable-widths consider ideal scenarios with specific bit-precision for MAC/memory for each layer which is not only impractical in real hardware but also is difficult to scale across different model architectures. 
Using this hardware platform, we perform a realistic energy cost analysis and show that our proposed method performs much better when compared to baseline models- with homogeneous precision for all layers. 

Additionally, we also perform an analytical energy estimation based on multiply-and-accumulate (MAC) and simple memory access calculations similar to that of many existing works in the literature \cite{han2015deep, QUANOS, PCA}. We do this to highlight the fact that such kind of analysis assumes impractical hardware architecture design scenarios which tend to overestimate the efficiency improvements, especially for mixed-precision networks. In this paper we make the following contributions:
\begin{itemize}
    \item We propose a novel in-training method to find optimal quantization bit-width for each layer based on \textit{Activation Density (AD)} without significant reduction in accuracy.
    \item We empirically evaluate our approach on CIFAR-10, CIFAR-100 and TinyImagenet datasets.
    \item As quantization is orthogonal to pruning, we also evaluate the performance of our algorithm when used simultaneously with Activation Density based pruning method.
    \item We design a PIM accelerator with in-built support for multi-bit data precision scalability, and perform realistic energy analysis comparison between quantized, quantized-and-pruned and baseline architectures.
    \item We show that our PIM hardware based energy analysis gives more realistic energy estimates than analytical energy analysis which often relies on impractical hardware architecture designs for mixed-precision models.
\end{itemize}

\section{Background}
\subsection{Quantization}
Quantization is the process of using a lower bit representation of activations and/or weights to achieve efficient computation. A 32-bit floating-point arithmetic (FP32) is the default in most of the modern deep learning implementations. However, numerous techniques have demonstrated that a similar accuracy can be achieved with much lower bit-widths such as 4-/2-/1-bit \cite{qnn}. Moreover, integer arithmetic is more efficient than floating point arithmetic lowering the energy budget significantly. In the cases of extreme quantization where there is 1-bit representation, the integer arithmetic can be further reduced to bit-wise XNOR operations \cite{xnornets}. 

Typical $k$-bit quantization is given by-
\small
\begin{equation}
x_q = round((x - x_{min})(\frac{2^k - 1}{x_{max} - x_{min}}))
\label{eqn:vanilla_quantization}
\end{equation}
\normalsize
where, $x$ denote original values and $x_q$ represent the corresponding quantized values. $x$ can be floating point representation or quantized representation with $\ge k$ quantization levels.

\subsection{Pruning}
Neural networks are known to have ample redundancies. The objective of pruning is to remove redundant connections in the neural network to make the model sparse. Pruning helps in reducing the model size that in turn reduces the storage and communication requirements for the models which is particularly useful in embedded devices for distributed intelligence. Pruning coupled with accelerators that can efficiently compute sparse matrix multiplications yield significant compute and energy efficiency \cite{han2015deep, weightsandconnections_han, parashar2017scnn}. 

\subsection{Activation Density}\label{AD_method}
\begin{figure}[h!]
    \centering
    \includegraphics[width=0.6\columnwidth]{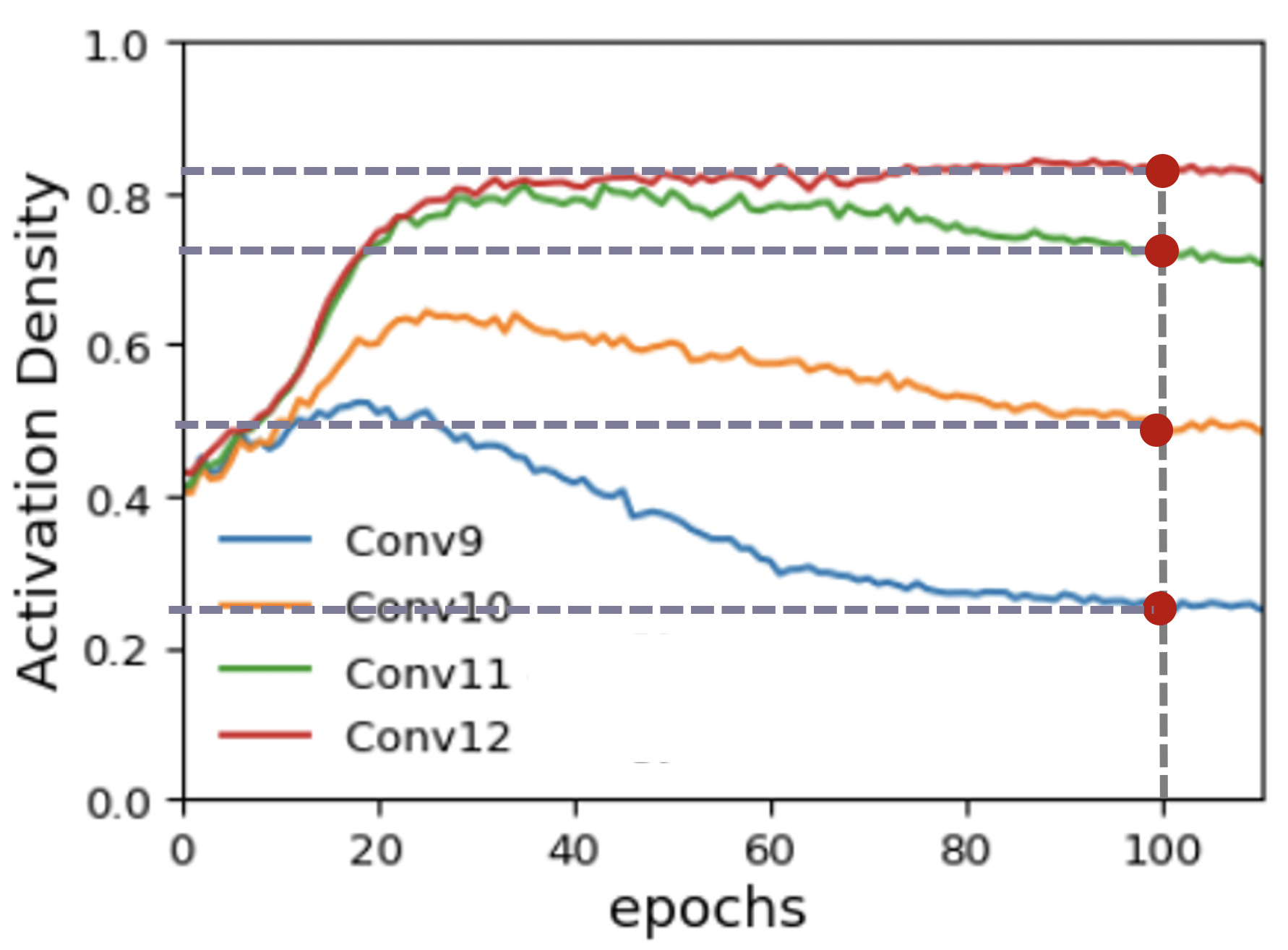}
    \caption{Trend of Activation Density (AD) of a few individual layers. Here, we observe that AD stabilizes at around 100 epochs. We use this observation to formulate a quantization method based on AD (see Algorithm 1).}
    \label{fig:ad_plot}
\end{figure}
In this work, we use a metric called Activation Density (AD) \cite{foldy2020activation} to determine the optimal bit-precision per layer. AD is defined as the proportion of non-zero activations in a layer calculated by passing the training set through the network-

\small
\begin{equation}\label{eq:AD}
    AD = \frac{\# nonzero\hspace{2mm} activations}{\#total\hspace{2mm}activations}
\end{equation}
\normalsize

Intuitively, AD quantifies the utilization of parameters in a network. For instance, for a layer with 512 neurons and 100 neurons yielding non-zero output, AD will be $100/512 = 0.195$. AD can also be calculated for the entire network by accumulating the statistics of all the layers. A noteworthy observation that forms the basis of our work is that the AD of the network saturates (to a value $<1$) as the training progresses (see Fig. 1). This implies that a majority of activations are rendered inactive during training, hinting that there is a huge redundancy in the model. We take advantage of this fact to quantize each layer of a network to lower bit-precisions based on AD. Thus, quantizing based on AD can be interpreted as removing the redundancies of a model by reducing the precision. Note, most deep learning  networks today use rectified linear unit (ReLU) activation functions which will generate a good proportion of zero/non-zero activations. Hence, AD can be used as a universal metric for quantifying redundancy. 

\section{Methodology}

Given a model $M$ with layers $l = 1, 2, ... n$, our approach aims to optimize quantization bit-width $k_l$ for each layer $l$ based on the AD metric. Given $A_l$ as activation output from layer $l$, the quantized activation $A_{lq}$, is calculated according to eqn. \ref{eqn:vanilla_quantization}. Similarly, during backpropagation after updating the weights $W$, the quantized weights $W_q$ are calculated as per eqn. \ref{eqn:vanilla_quantization}. The model is trained using standard backpropagation algorithm. Note, depending upon the quantization bit-width per layer, we use quantized activations and weights $A_q, W_q$ during forward propagation. During backpropagation, the gradients are calculated and the weights $W_q$ are updated. The updated weights are again quantized before the next training step. 
During training, we monitor the activation density $AD_l$ for all the layers. Once $AD_l$ stabilizes across all layers, we break the training process and then, perform quantization. The quantization bit-width $k_l$ for each layer $l$ is calculated as-

\small
\begin{equation}\label{eq:adq}
     k_l = round(k_{l_{initial}}*AD_l) 
\end{equation} 
\normalsize

For instance, in Fig. \ref{fig:ad_plot}, we observe that AD is nearly constant after $100$ epochs. In such case, we stop training at the 100th epoch, quantize the network using eqn. \ref{eq:adq} and then continue the training of the mixed-precision $k_l$-bit network. Say, $AD_l$ values of a model are \{0.9, 0.3, 0.5\} and the initial bit-widths are \{16, 10, 8\}, applying eqn. \ref{eq:adq} will yield bit-widths of \{14-bit, 3-bit, 4-bit\}, respectively. Note, both the weights and the activations of layer $l$ are quantized to $k_l$-bits.
The updated $k_l$-bit model is then trained for next set of epochs and this process is repeated till we no longer observe a change in $AD_l$ for any layer. In practice, we observe that the AD of the layers increases with each quantization iteration reaching the maximum limit $1.0$ when further quantization is not possible. The final $k_l$-bit model obtained when $AD_l=1$ is then trained till convergence. Generally, we find that if we start from a 16-bit model, our method converges to the final mixed-precision model within 3 to 4 iterations ($iter$ in Algorithm 1). Since we train progressively lower-precision models during the training process, we see that the overall training complexity reduces. Our method is summarized in Algorithm \ref{algorithm:main_algo}.

\begin{algorithm}[ht]\label{algorithm:main_algo}
\SetAlgoLined
 \textbf{Input}: Model $M$ with layers $l = 1, 2, ..., L$\;
 \textbf{Output}: Optimal bit width $k_l$, $\forall$ $l$ $\epsilon$ $M$\;
 Initialize model $M$ with random weights\;
 Set bit width $k_l^{(0)} = 16$ of initial model, $\forall$ $l$ $\epsilon$ $M$\;
 \For{iter = $1$ to $N$}{
    \For{epoch = $1$ to $\#(epochs)$}{
        Forward and Backward Propagation of $M$\;
        
        \tcp{Monitor Activation Density. Break if AD is saturated for all layers}   
        Compute $AD_l$ $\forall$ $l$ $\epsilon$ $M$ using eqn. \ref{eq:AD}\;
        \If{$AD_l$ is saturated $\forall$ $l$ $\epsilon$ $M$}{
            break\;
        }
     }
    \For{each layer $l$ in $M$}{
        \tcp{Calculate the new bit width using eqn. \ref{eq:adq}}
        $k_l^{(iter)} = round(k_l^{(iter-1)}*AD_l)$ $\forall$ $l$ $\epsilon$ $M$\;
    }
 }
 
 \caption{Activation Density Based Quantization}

\end{algorithm}
\vspace{-4mm} 

\begin{figure}[h]
    \centering
    \includegraphics[width=0.7\columnwidth]{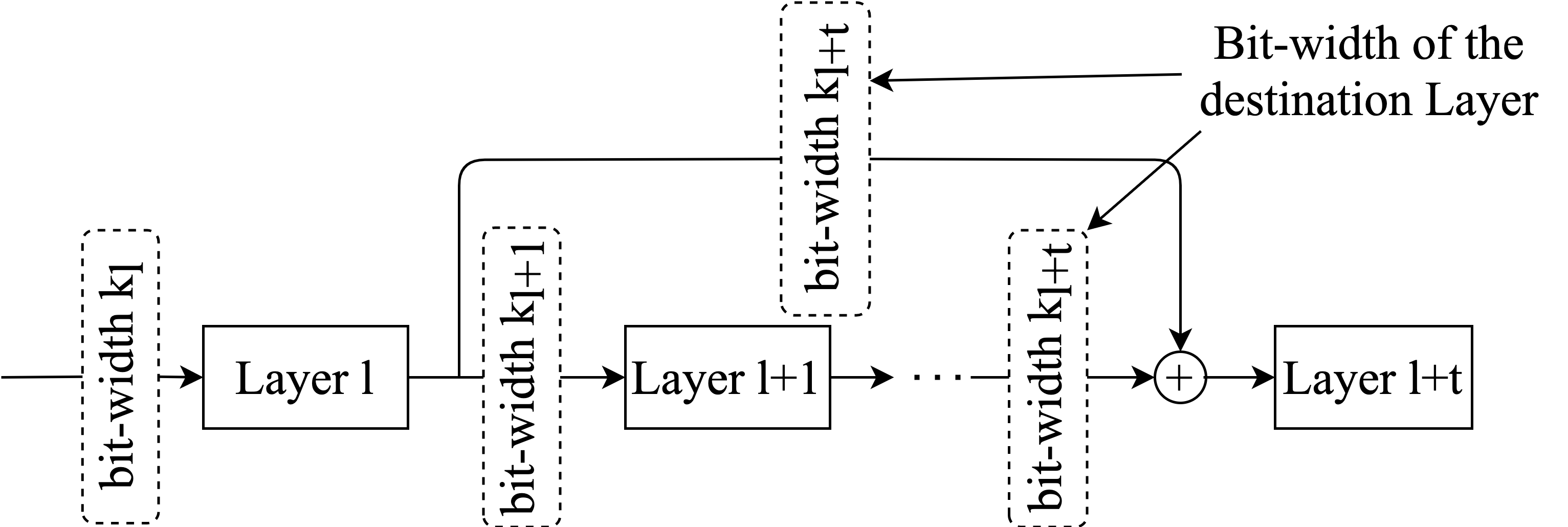}
    \caption{Illustration to show the bit-width calculation in skip connections.}
    \label{fig:resnet_quant}
    \vspace{-3mm}
\end{figure}

Note that for ResNet models, where there are skip connections between layers, we use the bit-width of the destination layer to quantize the activations of the skip branches as illustrated in Fig. \ref{fig:resnet_quant}. 

\section{Experiments and Results}
To validate our method, we run a series of experiments measuring accuracy, energy (section \ref{energy_estimation}) and training complexity (section \ref{training_complexity}) on CIFAR-10, CIFAR-100 and TinyImagenet datasets. The model is trained using Adam optimizer under standard settings. We illustrate the fundamentals of our approach by analysing the AD across all layers of a VGG19 model trained on CIFAR-10 before and after quantization in Fig \ref{fig:ad_vgg19_cifar10_adam_baseline} and \ref{fig:ad_vgg19_cifar10_adam_quantized}. We summarize our results for all the datasets in Table \ref{table:adq_summary}, \ref{table:adq_adp_summary}. Further, we present a hardware evaluation based on a PIM based accelerator in section \ref{hardware_evaluation}.

Fig. \ref{fig:ad_vgg19_cifar10_adam_baseline} shows the training progress along with the trend of AD for each layer of a 16-bit baseline VGG19 model across different training epochs. We can observe that AD converges to a value $<1.0$ for all layers implying that there is a considerable amount of redundancy in the baseline model. In contrast, as observed in Fig. \ref{fig:ad_vgg19_cifar10_adam_quantized}, when AD based quantization is applied as per Algorithm 1, AD reaches $\simsym1.0$ implying an adequate utilization of layers. Interestingly, the AD of the last layer is very low in spite of extreme quantization (bit-width of $1$) suggesting that we can entirely remove that layer. This is validated by the results in Table \ref{table:adq_summary} as the network in \textit{iteration 2a} which has its last layer removed achieves similar accuracy compared to network at \textit{iteration 2}. Note that in Fig. \ref{fig:ad_vgg19_cifar10_adam_baseline} we show the plots upto full $210$ epochs of training in order to observe the AD pattern. In practice, we perform quantization and go to the subsequent iteration of training the $k_l$-bit model once the AD of the previous model stabilizes. For Fig. \ref{fig:ad_vgg19_cifar10_adam_baseline}, we essentially stop training the 16-bit baseline at 100th epoch when $AD_l$ across all layers has almost saturated.  Tables \ref{table:adq_summary} and \ref{table:adq_adp_summary} summarize all the useful metrics of our method. We specify the bit-width of each layer achieved by our method for each layer in the architecture. Note that for ResNet18, we omit the bit-precision and number of channels for convolutional layers in the skip connections as they are equal to that of the destination layer of the skip connection. Also, across all our experiments, we do not quantize the first layer and the final fully connected layer to avoid a drastic drop in accuracy. We report the  accuracy on the test dataset, the overall AD averaged across all layers of the corresponding $k_l$-bit model, number of epochs trained in each quantization iteration (or $iter$ in Algorithm 1), energy efficiency and training complexity. 
\begin{figure}[t]
    \centering
    \includegraphics[width=0.85\columnwidth]{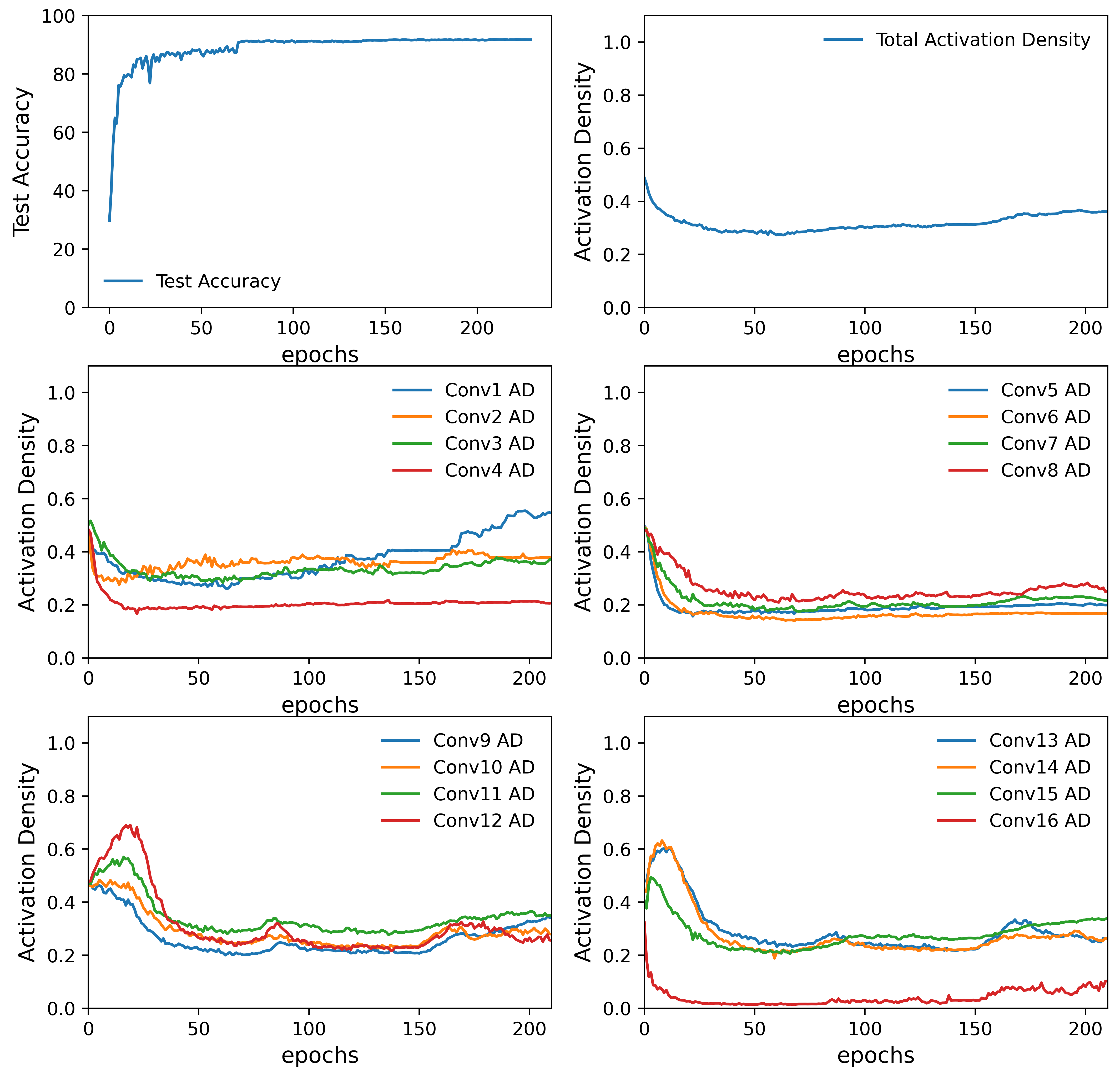}
    \caption{Trend of Accuracy and Activation Density (AD) vs epochs for VGG19 layers on CIFAR-10: Baseline (Table \ref{table:adq_summary} (a), Iter 1).}
    \label{fig:ad_vgg19_cifar10_adam_baseline}
    \vspace{-4mm}
\end{figure}
\begin{figure}[h]
    \centering
    \includegraphics[width=0.85\columnwidth]{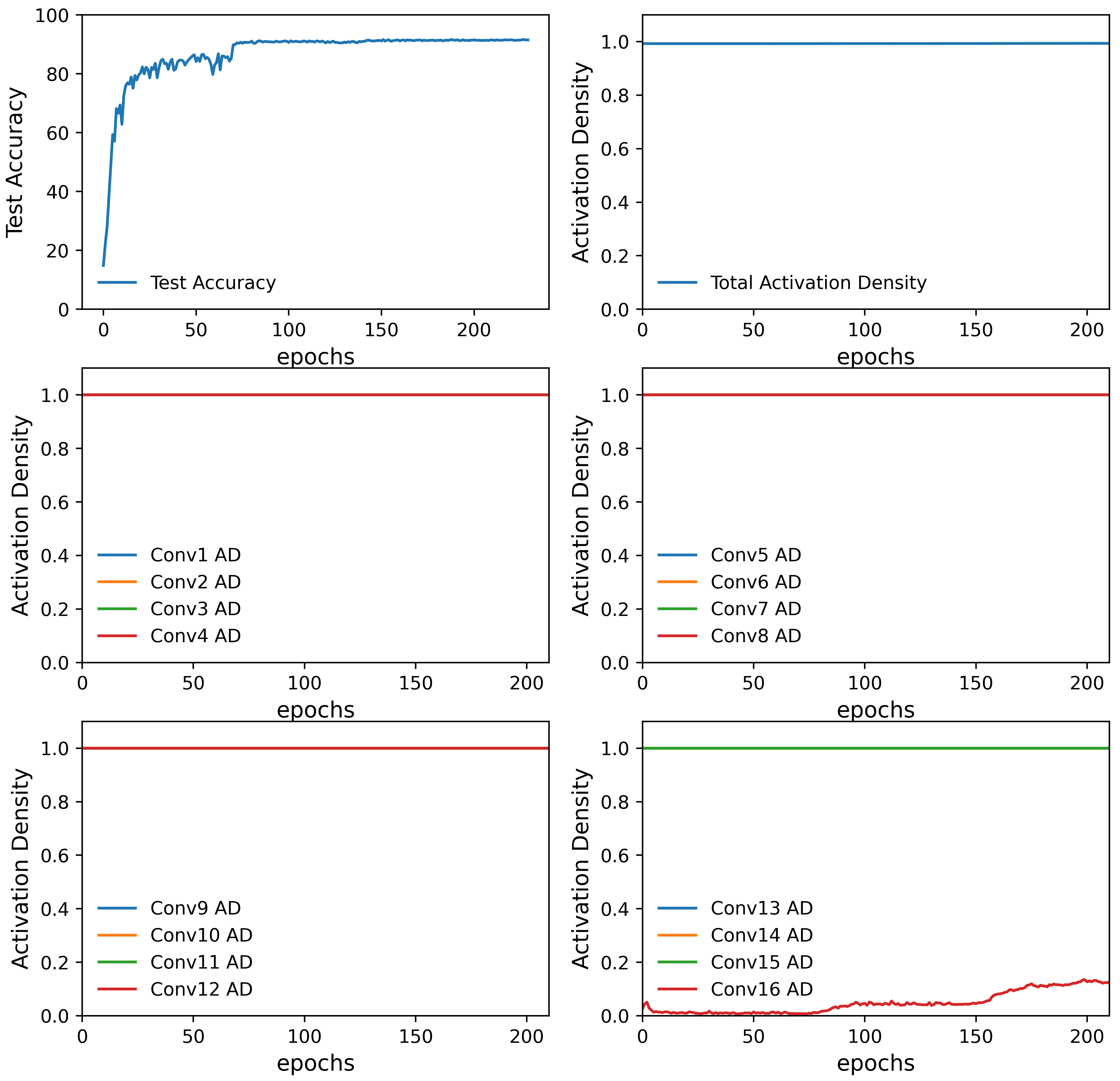}
    \caption{Trend of Accuracy and Activation Density (AD) vs epochs for VGG19 layers on CIFAR-10: With Activation Density based Quantization (Table \ref{table:adq_summary} (a), Iter 2). In contrast to fig \ref{fig:ad_vgg19_cifar10_adam_baseline}, here we can observe that AD is close to 1 in most of the layers implying maximum utilization.}
    \label{fig:ad_vgg19_cifar10_adam_quantized}
    \vspace{-4mm}
\end{figure}
\subsection{Analytical Energy Estimation}\label{energy_estimation}
We analytically approximate the energy consumption on traditional hardware based on 45 nm CMOS process using estimates described in \cite{weightsandconnections_han, QUANOS}. This analytical energy analysis is done for two reasons: 1) To get a rough estimate of the energy consumption without any need for hardware setup, 2) To highlight the contrast between such analytical and real hardware energy estimations (as will be shown in later section V). Note that, for analytical estimations, we are considering only MAC and memory access operations and neglecting any peripheral circuit energy. We provide a comprehensive evaluation considering all overheads on PIM hardware in section \ref{hardware_evaluation}. The energy for $k_l$-bit memory access and MAC operations are summarized in Table \ref{tab:energy_calc}.
For a $k_l$-bit $p \times p$ convolution layer with $I$ input channels, $O$ output channels operating on an input feature map size $N \times N$ resulting in output feature map of size $M \times M$, the number of memory accesses ($N_{mem}$), number of MAC operations ($N_{MAC}$) and total energy per layer $l$ ($E_l$) is given by: $ N_{Mem} = N^2 \times I + p^2 \times I \times O$, $N_{MAC} = M^2 \times I \times p^2 \times O$ and $E_l = N_{Mem} * E_{Mem|k_l} + N_{MAC} * E_{MAC|k_l}$ respectively. Note that if all the weights and activations become binary this method does not hold as the compute reduces to XNOR and popcount.

\begin{table}[t]
    \centering
    \caption{Energy Consumption Estimates}
    \resizebox{0.7\linewidth}{!}{
    \begin{tabular}{|c|c|}
    \hline
    \textbf{Operation} & \textbf{Estimated Energy (pJ)} \\
    \hline
    $k_l$-bit Memory access ($E_{Mem|k_l}$) & $2.5k_l$  \\
    32-bit Multiply ($E_{Mult|32}$) & 3.1\\
    32-bit Add ($E_{Add|32}$) & 0.1 \\
    $k_l$-bit Multiply and Accumulate ($E_{MAC|k_l}$) & $((3.1 * k_l)/32 + 0.1)$\\
    \hline
    \end{tabular}
    }
    \label{tab:energy_calc}
    \vspace{-5mm}
\end{table}


From Table \ref{table:adq_summary}, we see our method achieves energy efficiency of $4.19 \times$ in VGG19 model trained on CIFAR-10 dataset with no drop in accuracy. Further, with ResNet18 network trained on CIFAR-100 and TinyImagenet datasets, we obtain $3.19\times$ and $4.5\times$ energy efficiency, respectively, at near iso-accuracy with baseline 16-bit precision model. 

\subsection{Training Complexity}\label{training_complexity}
Since we are iteratively quantizing the model during training, we considerably reduce the amount of time and compute required to train  the model. To calculate the savings in terms of training time and energy, we use the \textit{Training Complexity} metric \cite{foldy2020activation} defined as:

\small
\begin{equation}\label{eqn:training_complexity}
    \sum_{iter=i}^{N} (\text{MAC reduction}_{i})^{-1} \times (\text{\# epochs}_{i})
\end{equation}
\normalsize

where, $i$ is the denotes the quantization iteration $iter$ from Algorithm \ref{algorithm:main_algo}, $\# epochs_i$ is the number of epochs that the model is trained in each iteration. In Table \ref{table:adq_summary}, we observe a significant $50\%$ reduction in training complexity in VGG19 model trained on CIFAR-10. We observe a similar improvement in training complexity for CIFAR-100 and TinyImagenet as well.

\subsection{Activation Density based Pruning}
In addition to quantization, we also evaluate our method when used simultaneously with AD based pruning proposed in \cite{foldy2020activation}. Thus, along with quantization, we iteratively prune the channels of each layer of a network based on its AD as

\small
\begin{equation}
    C_l = round(C_{l_{initial}}*AD_l)
    \vspace{-1mm}
\end{equation}
\normalsize

where, $C_l$ and $AD_l$ are the number of channels and the activation density of layer $l$, respectively.

When used together, AD based pruning and quantization achieve $980\times$ reduction in estimated energy with $<5\%$ drop in accuracy with VGG19 model trained on CIFAR-10 dataset. Similarly for ResNet18 on CIFAR-100 and TinyImagenet datasets, we observe energy efficiency ranging from $100\times$ to $300\times$ with $\simsym5\%$ accuracy drop. Using pruning with quantization also translates to improvement in training complexity (not shown) since there will be less number of channels as our method progresses.

\begin{table*}[t]
    \centering
        \caption{Results Summary: Activation Density based Quantization }
    \resizebox{0.8\linewidth}{!}{
    \begin{tabu} to \textwidth{|X[0.5]|X[7.5]|X[0.5]|X[0.5]|X[0.5]|X[0.5]|X[0.5]|}
        
        \hline
        \textbf{Iter} & \textbf{Architecture (Layer-wise bit-width)} & \textbf{Test Accuracy} & \textbf{Total AD} & \textbf{Energy Efficiency} & \textbf{No. of Epochs} & \textbf{Train Compl.} \\
        \hline
        \multicolumn{7}{|c|}{(a) VGG19 on CIFAR-10}\\
        \hline
        1 & 16-bit Full Precision for all layers & 91.85\% & 0.284 & 1x & 100 & 1x \\
        \hline
        
        2 & bit-width: [16, 4, 5, 4, 3, 2, 2, 2, 3, 3, 3, 4, 3, 3, 3, 3, 16] & 91.62\% & 0.992 & 4.16x & 70 & 0.524x \\
        \hline
        2a & bit-width: [16, 4, 5, 4, 3, 2, 2, 2, 3, 3, 3, 4, 3, 3, 3, x, 16] & 92.16\% & 1.000 & 4.19x & 70 & 0.502x \\
        \hline
        
        \multicolumn{7}{|c|}{(b) ResNet18 on CIFAR-100}\\
        
        
        \hline
        1 & 16-bit Full Precision for all layers & 70.90\% & 0.416 & 1x & 120 & 1x \\
        \hline
        2 & bit-width: [16, 5, 3, 3, 11, 1, 1, 11, 4, 4, 10, 4, 4, 11, 3, 3, 9, 3, 3, 9, 3, 3, 6, 1, 1, 16] & 71.51\% & 0.743 & 2.76x & 70 & 0.620x \\
        \hline
        3 & bit-width: [16, 5, 3, 3, 5, 1, 1, 8, 4, 4, 6, 4, 4, 8, 3, 3, 9, 3, 3, 9, 3, 3, 6, 1, 1, 16] & 70.51\% & 0.869 & 3.19x & 70 & 0.703x \\
        \hline
        
        \multicolumn{7}{|c|}{(c) ResNet18 on TinyImagenet}\\

        
        \hline
        1 & 32-bit Full Precision for all layers & 44.26\% & 0.447 & 1x & 60 & 1x \\
        \hline
        2 & bit-width: [16, 10, 7, 7, 22, 10, 10, 24, 10, 10, 22, 6, 6, 22, 9, 9, 18, 5, 5, 16, 4, 4, 11, 3, 3, 16] & 43.94\% & 0.651 & 2.73x & 25 & 0.694x \\
        \hline
        3 & bit-width: [16, 3, 7, 7, 16, 2, 2, 17, 3, 3, 15, 6, 6, 15, 9, 9, 9, 5, 5, 7, 4, 4, 4, 3, 3, 16] & 44.00\% & 0.914 & 4.14x & 25 & 0.705x \\
        \hline
        4 & bit-width: [16, 3, 7, 7, 14, 2, 2, 14, 3, 3, 10, 6, 6, 10, 9, 9, 9, 5, 5, 7, 4, 4, 4, 3, 3, 16] & 43.50\% & 0.917 & 4.50x & 25 & 0.770x \\
        \hline
    \end{tabu}
    }
    \label{table:adq_summary}
    \vspace{-4mm}
\end{table*}

\begin{table*}[t]
    \centering
        \caption{Results Summary: Activation Density based Quantization coupled with Activation Density based Pruning}
    \resizebox{0.8\linewidth}{!}{
    \begin{tabu} to \textwidth{|X[0.5]|X[7.5]|X[0.5]|X[0.5]|X[0.5]|X[0.5]|X[0.5]|}

        \hline
        \textbf{Iter} & \textbf{Architecture (Layer-wise bit-width and number of channels)} & \textbf{Test Accuracy} & \textbf{Total AD} & \textbf{Energy Efficiency} & \textbf{No. of Epochs} & \textbf{Train Compl.} \\
        \hline
        \multicolumn{7}{|c|}{(a) VGG19 on CIFAR-10}\\
        \hline
        \multirow{2}{*}{1} & 16-bit Full Precision for all layers & \multirow{2}{*}{91.85\%} & \multirow{2}{*}{0.284} & \multirow{2}{*}{1x} & \multirow{2}{*}{100} & \multirow{2}{*}{1x} \\
        \cline{2-2}
         & nchannels: [64, 64, 128, 128, 256, 256, 256, 256, 512, 512, 512, 512, 512, 512, 512, 512] & & & & & \\
        \hline
        \multirow{2}{*}{2} & bit-width: [16, 4, 5, 9, 4, 3, 5, 2, 2, 2, 3, 5, 3, 3, 4, 3, 4, 3, 3, 3, 16] & \multirow{2}{*}{86.88\%} & \multirow{2}{*}{0.999} & \multirow{2}{*}{980x} & \multirow{2}{*}{70} & \multirow{2}{*}{0.344x} \\
        \cline{2-2}
         & nchannels: [19, 22, 38, 24, 45, 37, 44, 54, 103, 126, 150, 125, 122, 112, 111, 8] & & & & & \\
        \hline
        
        \multicolumn{7}{|c|}{(b) ResNet18 on CIFAR-100}\\
        
        \hline
        \multirow{2}{*}{1} & 16-bit Full Precision for all layers & \multirow{2}{*}{70.90\%} & \multirow{2}{*}{0.416} & \multirow{2}{*}{1x} & \multirow{2}{*}{120} & \multirow{2}{*}{1x} \\
        \cline{2-2}
         & nchannels: [64, 64, 64, 64, 64, 128, 128, 128, 128, 256, 256, 256, 256, 512, 512, 512, 512] & & & & & \\
        \hline
        \multirow{2}{*}{2} & bit-width: [16, 5, 3, 11, 1, 11, 4, 10, 4, 11, 3, 9, 3, 9, 3, 6, 1, 16] & \multirow{2}{*}{66.40\%} & \multirow{2}{*}{0.732} & \multirow{2}{*}{150x} & \multirow{2}{*}{70} & \multirow{2}{*}{0.372x} \\
        \cline{2-2}
         & nchannels: [21, 12, 44, 6, 47, 34, 87, 34, 89, 58, 156, 50, 146, 110, 192, 59, 59] & & & & & \\
        \hline
        \multirow{2}{*}{3} & bit-width: [16, 5, 3, 5, 1, 8, 4, 6, 4, 8, 3, 9, 3, 9, 3, 6, 1, 16] & \multirow{2}{*}{63.01\%} & \multirow{2}{*}{0.992} & \multirow{2}{*}{300x} & \multirow{2}{*}{70} & \multirow{2}{*}{0.374x} \\
        \cline{2-2}
         & nchannels: [21, 12, 19, 1, 31, 34, 61, 34, 58, 58, 156, 50, 146, 110, 192, 9, 22] & & & & & \\
        \hline
        
        \multicolumn{7}{|c|}{(c) ResNet18 on TinyImagenet}\\
        
        \hline
        \multirow{2}{*}{1} & 32-bit Full Precision for all layers & \multirow{2}{*}{44.26\%} & \multirow{2}{*}{0.447} & \multirow{2}{*}{1x} & \multirow{2}{*}{60} & \multirow{2}{*}{1x} \\
        \cline{2-2}
         & nchannels: [64, 64, 64, 64, 64, 128, 128, 128, 128, 256, 256, 256, 256, 512, 512, 512, 512] & & & & & \\
        \hline
        \multirow{2}{*}{2} & bit-width: [16, 10, 7, 22, 10, 24, 10, 22, 6, 22, 9, 18, 5, 16, 4, 11, 3, 16] & \multirow{2}{*}{38.40\%} & \multirow{2}{*}{0.666} & \multirow{2}{*}{93.4x} & \multirow{2}{*}{25} & \multirow{2}{*}{0.450x} \\
        \cline{2-2}
         & nchannels: [20, 14, 45, 21, 48, 42, 88, 27, 91, 73, 151, 41, 129, 70, 178, 56, 20]  & & & & & \\
        \hline
        
    \end{tabu}
    }
    \label{table:adq_adp_summary}
    \vspace{-3mm}
\end{table*}

\begin{table}[t]
\centering
\caption{Energy values for a single MAC operation for multiple bit-precisions using our proposed PIM accelerator}
\label{tab:energy_val}
\resizebox{0.6\linewidth}{!}{
\begin{tabular}{|c|c|}
\hline
\textbf{Energy due to a MAC operation} & \textbf{Value (fJ)} \\ \hline
$E_{MAC|2-bit}$ & 2.942 \\ \hline
$E_{MAC|4-bit}$ & 16.968 \\ \hline
$E_{MAC|8-bit}$ & 66.714 \\ \hline
$E_{MAC|16-bit}$ & 276.676 \\ \hline
\end{tabular}
}
\vspace{-4mm}
\end{table}

\begin{table}[t]
\centering
\caption{Comparison of PIM hardware MAC energy of mixed-precision model with baseline- unpruned full-precision model}
\label{tab:compare-mac}
\resizebox{0.9\linewidth}{!}{
\begin{tabular}{|p{2.5cm}|p{2.5cm}|p{2.5cm}|p{2cm}|}
\hline
\textbf{Network \& Dataset} & \textbf{Energy consumed by mixed-precision network ($\mu J$)} & \textbf{Energy consumed by full-precision network ($\mu J$)} & \textbf{Energy reduction} \\ \hline
VGG19 on CIFAR-10 dataset & 21.506 & 110.154 & 5.12x \\ \hline
ResNet18 on CIFAR-100 dataset & 33.186 & 159.501 & 4.81x \\ \hline
\end{tabular}
}
\vspace{-6mm}
\end{table}

\begin{table}[t]
\centering
\caption{PIM hardware MAC energy of mixed-precision model (with pruning) Vs baseline- unpruned full-precision model}
\label{tab:compare-mac-pruning}
\resizebox{0.9\linewidth}{!}{
\begin{tabular}{|p{2.5cm}|p{2.5cm}|p{2.5cm}|p{2cm}|}
\hline
\textbf{Network \& Dataset} & \textbf{Energy consumed by pruned mixed-precision network ($\mu J$)} & \textbf{Energy consumed by full-precision network ($\mu J$)} & \textbf{Energy reduction} \\ \hline
VGG19 on CIFAR-10 dataset & 0.558 & 110.154 & 197.55x \\ \hline
ResNet18 on CIFAR-100 dataset & 3.630 & 159.501 & 43.941x \\ \hline
\end{tabular}
}
\vspace{-4mm}
\end{table}

\section{PIM Hardware Evaluation}\label{hardware_evaluation}
\begin{figure}[t]
    \centering
    \includegraphics[width=0.3\textwidth]{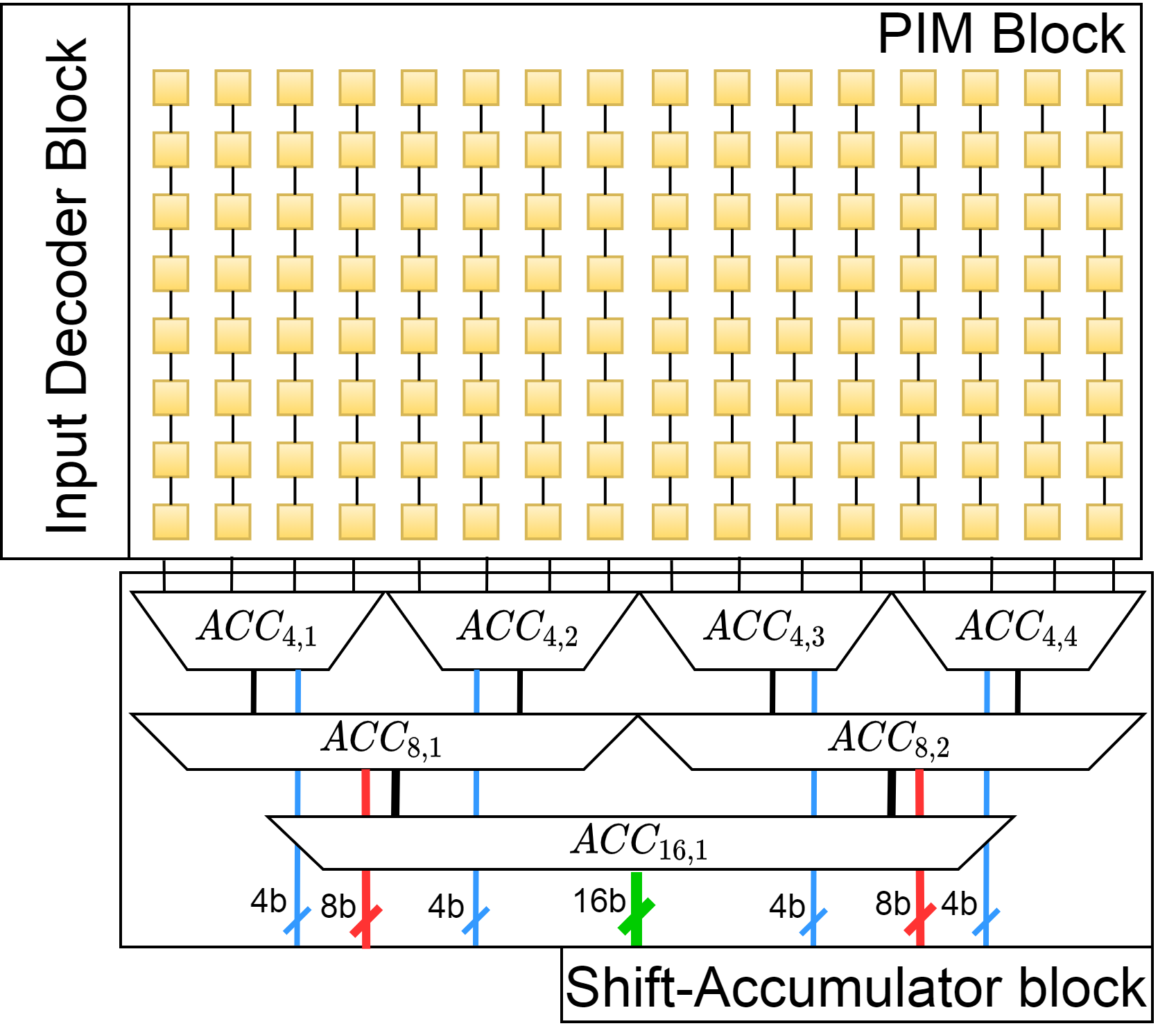}
    \caption{PIM hardware accelerator with shift-accumulators for mixed-precision support}
    \label{acce_arch}
    \vspace{-4mm}
\end{figure}
\subsection{PIM Hardware Setup}

To evaluate the working of our proposed method, we develop a Process In-Memory (PIM) accelerator hardware which can support multi-bit precision for various layers of a network \cite{qpim}. The accelerator architecture shown in Fig. \ref{acce_arch} has three main sections: 1) Input decoder, 2) PIM block 3) Shift-Accumulator block. The \textit{Input Decoder Block} fetches the activation values from layer $l-1$ and feeds them to the \textit{PIM block} of the layer $l$. This is done in a structured pattern. The \textit{PIM block} is a 2-D array of 1-bit SRAM-memory-and-multiply cells each performing 1-bit multiplication between the input activations and weights (stored inside the SRAM cells). After the multiplication, the \textit{Shift-Accumulator block}, which contains a series of accumulators, perform shift-and-add operations. The lowest level of the accumulator block in our PIM architecture is 4-bit. The next two levels perform 8-bit and 16-bit operations. Essentially, 4 columns of the PIM array are read together and the results are stored in the lowest block. Depending upon the bit-width of a given layer, each level of the \textit{Shift-Accumulator block} is activated. For example, if the weight/activation bit-width of a given layer is 2-bits, the corresponding MAC values are stored in the 4-bit accumulator {$ACC_{4,i}$} and are regarded as the final result and forwarded as shown using the blue lines. Similarly, if the precision of weights/activations are 4-bits, the results from {$ACC_{4,i}$} accumulators undergo shift-and-add to yield 8-bit accumulated results in {$ACC_{8,i}$} which are then forwarded, shown using red lines.

\subsection{Energy Evaluation}
In a PIM architecture, energy is primarily expended during MAC operation as memory access energy is greatly reduced. Also, energy due to peripheral components is fairly minimal and have not been considered for evaluation. In our architecture, energy is consumed by the \textit{PIM block} and the \textit{Shift-Accumulator block}. Table~\ref{tab:energy_val} lists the energy consumed for a single MAC operation for multiple bit-precisions evaluated on 45nm CMOS. Using this, we compute the total energy consumption for our model with AD-based quantization/pruning and compare the results with baseline 16-bit full precision models. Table~\ref{tab:compare-mac} reports the energy comparison for VGG19, Reset18 networks on CIFAR-10, CIFAR-100 data with and without AD-based-quantization. We obtain $\sim5\times$ higher energy efficiency over baseline models for both the networks. Similarly, when AD-based quantization-and-pruning is simultaneously applied on the model, improvements of $\sim197\times$ and $\sim43\times$ for VGG19 and ResNet18 networks, respectively, are observed (Table~\ref{tab:compare-mac-pruning}).

As discussed in Section~\ref{sec:intro}, we perform a realistic energy analysis pertaining to more practical hardware architecture considerations as opposed to the analytical estimates shown in Section~\ref{energy_estimation}. Thus, during analytical estimations in Table \ref{table:adq_adp_summary}, we get overestimated energy efficiencies $\sim5-7\times$ greater than practical hardware implementations (Table \ref{tab:compare-mac-pruning}).


\section{Conclusion}

We present a simple in-training quantization method based on Activation Density (AD) that enables us to compute the optimal bit-precision of each layer of a network during training. Our approach yields an energy-efficient mixed-precision model with iso-accuracy with baseline. We evaluate the effectiveness of our method on VGG19 and ResNet18 neural networks with multiple benchmark datasets and report our findings. We find that the AD-based quantization approach reduces the training complexity by $50\%$ in our experiments alongside providing up to $4.5$x benefit with respect to OPS reductions. We also assess the performance of our algorithm when used simultaneously with AD-based pruning during training. Finally, we propose a scalable PIM accelerator that supports multiple bit-precisions and pruning for different layers of a network. We perform a realistic energy cost analysis to show that our proposed AD-based quantization/pruning algorithm achieves significant energy efficiency improvements $\sim5 -100\times$ when compared with baseline 16-bit models.  

\section*{Acknowledgement}
This work was supported in part by the National Science Foundation, Amazon Research Award, and Technology Innovation Institute Abu Dhabi.
\vspace{-2.5mm}
\bibliographystyle{IEEEtran}
\bibliography{bib}

\end{document}